\documentclass[accepted]{uai2023} 

\usepackage[american]{babel}

\usepackage{natbib} 
\usepackage{mathtools} 
\usepackage{booktabs} 
\usepackage{tikz} 

\usepackage{hyperref}       
\usepackage{microtype}      
\usepackage{graphicx}
\usepackage{url}            
\usepackage{amsfonts}       
\usepackage{nicefrac}       
\usepackage{lipsum}
\usepackage{amsmath}
\usepackage{subcaption}     
\usepackage{enumitem}
\usepackage{wrapfig}
\usepackage{algorithmic}
\usepackage{amssymb}
\usepackage{epstopdf}
\usepackage{bbold}
\usepackage{xcolor}
\usepackage{algorithm}
\usepackage{multirow}
\usepackage{amsthm}
\usepackage{marginnote}
\usepackage{xfrac}
\usepackage{adjustbox}
\usepackage{colortbl}
\usepackage{tablefootnote}

\title{Fast FixMatch: Faster Semi-Supervised Learning with Curriculum Batch Size}

\author[1]{John Chen}
\author[1]{Chen Dun}
\author[1]{Anastasios Kyrillidis}
\affil[1]{%
    Computer Science Dept.\\
    Rice University\\
    Houston, Texas, USA
}

\begin{document}
\maketitle

\begin{abstract}
Advances in Semi-Supervised Learning (SSL) have almost entirely closed the gap between SSL and Supervised Learning at a fraction of the number of labels. 
However, recent performance improvements have often come \textit{at the cost of significantly increased training computation}. 
To address this, we propose Curriculum Batch Size (CBS), \textit{an unlabeled batch size curriculum which exploits the natural training dynamics of deep neural networks.} 
A small unlabeled batch size is used in the beginning of training and is gradually increased to the end of training. 
A fixed curriculum is used regardless of dataset, model or number of epochs, and reduced training computations is demonstrated on all settings. 
We apply CBS, strong labeled augmentation, Curriculum Pseudo Labeling (CPL) \citep{FlexMatch} to FixMatch \citep{FixMatch} and term the new SSL algorithm Fast FixMatch. 
We perform an ablation study to show that strong labeled augmentation and/or CPL do not significantly reduce training computations, but, in synergy with CBS, they achieve optimal performance. 
Fast FixMatch also achieves substantially higher data utilization compared to previous state-of-the-art.
Fast FixMatch achieves between $2.1\times$ - $3.4\times$ reduced training computations on CIFAR-10 with all but 40, 250 and 4000 labels removed, compared to vanilla FixMatch, while attaining the same cited state-of-the-art error rate \citep{FixMatch}. Similar results are achieved for CIFAR-100, SVHN and STL-10.
Finally, Fast MixMatch achieves between $2.6\times$ - $3.3\times$ reduced training computations in federated SSL tasks and online/streaming learning SSL tasks, which further demonstrate the generializbility of Fast MixMatch to different scenarios and tasks.
\end{abstract}

\section{Introduction}
\textcolor{purple}{\textbf{Background on SSL.}}
Semi-Supervised Learning (SSL) has shown remarkable results in the past few years and has increasing significance due to the plethora of unlabeled data \citep{FixMatch, FlexMatch, LargeScaleSSL1, LargeScaleSSL2}. 
Oftentimes, for Computer Vision, Natural Language Processing and other applications, there exist significant amounts of unlabeled data, either self-generated, open-source, or on the internet \citep{imagenet, LargeScaleData1, LargeScaleData2}. 
Obtaining labels in a large scale fashion is often difficult, due to monetary, scarcity, privacy or other reasons. 
As a result, there is a need to continuously improve the top-line performance of SSL algorithms.

FixMatch \citep{FixMatch} is a SSL algorithm which combines simplicity with performance. 
To set up the background for our discussion, consider the simple task of image classification based on the CIFAR-10 dataset.
Within each unlabeled data batch, FixMatch applies a pseudo-labeling technique to weakly augmented these unlabeled samples.
Then, it uses those labels to train with strongly augmented versions of the same unlabeled samples, based on the cross-entropy loss. 
FixMatch achieves exceptional results, such as 94.93\% accuracy on CIFAR-10 with 250 labels, and 88.61\% accuracy with only 40 labels.

\textcolor{purple}{\textbf{Challenges and potential solutions.}}
Yet, one of the main challenges of FixMatch is the computation required for training.
For example, it takes almost $8 \cdot 10^8$ forward and backwards passes to reach the 94.93\% accuracy on CIFAR-10 with 250 labels.
This is \textit{equivalent to performing approximately $15\cdot 10^3$ epochs} for the CIFAR-10 dataset, while state of the art accuracy for supervised learning on CIFAR-10 is achieved within a few hundreds of epochs.

One solution to this issue is by tackling the \textit{data utilization ratio}. 
In particular, by maintaining a fixed threshold throughout training, many forward passes of the weakly augmented samples never reach the pre-defined threshold and, therefore, never contribute to training. 
Previous work, such as FlexMatch \citep{FlexMatch}, attempted to address this issue by introducing a per-class dynamic threshold termed Curriculum Pseudo Label (CPL). 
However, while CPL increases data utilization and can improve final performance, it suffers a critical drawback: 
in early-mid training stages, a low threshold may mislead the model, and thus, it could be the case that \textit{CPL does not reduce total training computations.} 
Overall, while progress has been made on the data utilization problem, there is still room for improvement for the overarching challenge of reducing training computations.

\textcolor{purple}{\textbf{Improving the data utilization ration using CBS.}}
This work 
introduces Curriculum Batch Size (CBS). 
CBS applies a curriculum to the unlabeled batch size, which exploits the natural training dynamics of deep neural networks. 
A small unlabeled batch size is used at the beginning of training, and is gradually increased towards the end of training. 
\textit{A fixed curriculum is used regardless of dataset, model or number of epochs, and reduced training computations are demonstrated on all settings.}
We apply CBS, strong labeled augmentation, and CPL \citep{FlexMatch} to FixMatch \citep{FixMatch}, and term the new SSL algorithm Fast FixMatch. 
While strong labeled augmentations and/or CPL do not significantly reduce training computations at all times, they have synergy with CBS and together produce the best results.

\textcolor{purple}{\textbf{Summary of contributions and main observations.}} These are as follows: \vspace{-0.2cm}
\begin{itemize}[leftmargin=*]
    \item We propose the Curriculum Batch Size (CBS) schedule in the SSL setting. CBS introduces no extra hyperparameters to tune, and can be directly applied to FixMatch with no modifications across all tested settings. We apply CBS, strong labeled augmentation, and CPL \citep{FlexMatch} to FixMatch \citep{FixMatch}, and term the new SSL algorithm Fast FixMatch. \vspace{-0.1cm}
    \item We perform ablation studies to show that strong labeled augmentation and/or CPL do not significantly reduce training computations. Adding CBS to strong labeled augmentation and CPL together produces synergy and achieves optimal performance.
    \item We extend Fast FixMatch to Federated Self-supervised Learning scenario and online/streaming learning Self-supervised Learning scenario without introducing any extra parameters/computation cost.
\end{itemize}
We observe: \vspace{-0.2cm}
\begin{itemize}[leftmargin=*]
    \item CBS substantially increases the data utilization ratio, which allows more samples to be used effectively. In particular, CBS + CPL improves over CBS or CPL as much as CBS or CPL does over vanilla FixMatch. \vspace{-0.1cm}
    \item Fast FixMatch reduces training computations by $2.1\times$ - $3.4\times$ on CIFAR-10 with all but 40, 250 and 4000 labels removed, compared to vanilla FixMatch, while attaining the same cited state-of-the-art error rate. Similar results are achieved for CIFAR-100, SVHN and STL-10.
    \item Fast FixMatch reduces training computations by $2.6\times$ - $3.3\times$ on both CIFAR10 and CIFAR100 in Federated SSL with strong non-iid labeled and non-labeled local data. Similarly, Fast FixMatch reduces training computations by $2.3\times$ - $2.8\times$ on both CIFAR10 and CIFAR100 in online/streaming SSL with streaming of unlabeled data.
\end{itemize}

\section{Related Work}
\subsection{Semi-Supervised Learning}


\vspace{-0.2cm}
\textcolor{purple}{\textbf{Early Consistency Regularization.}}
Consistency regularization has been one of the main drivers for improvements in the SSL setting in recent years. Consistency regularization minimizes the difference between the outputs of augmentations of the same input, where defining the difference and choosing good augmentations have led to significant advances \citep{miyato2017virtual, berthelot2019mixmatch, FixMatch}. The $\pi$ model \citep{sajjadi2016regularization, laine2017temporal} adds a mean squared loss which minimizes differences on the output layer. Virtual Adversarial Training \citep{miyato2017virtual} perturbs the input with an adversarial perturbation by backpropagating the gradient to the input. The authors additionally add an entropy minimization \citep{grandvalet2005entmin} loss which encourages confident predictions for the top-1 class. Mean Teacher \citep{tarvainen2017mean} minimizes the difference between the current model and a model which is the exponential moving average of model weights.

\textcolor{purple}{\textbf{Strong Augmentation.}}
In the last few years, SSL have improved hand in hand with improved data augmentation. These include Mixup \citep{zhang2017mixup} which generates convex combinations of both inputs and labels, CutOut \citep{devries2017improved} which randomly removes parts of images, CutMix \citep{yun2019cutmix} which randomly replaces parts of an image with parts of another image and combining the label accordingly, AutoAugment \citep{cubuk2018autoaugment} which uses a reinforcement learning approach to generate the best data augmentations, RandAugment \citep{cubuk2020randaugment} which significantly speeds up AutoAugment, and many more \citep{hendrycks2020augmix, chen2022stackmix, fastautoaugment}.

MixMatch \citep{berthelot2019mixmatch} utilizes Mixup, temperature sharpening and other changes. Interpolation Consistency Training \citep{verma2019ict} uses Mixup and an exponential moving average of model weights. UDA \citep{xie2019unsupervised} demonstrates the effectiveness of RandAugment on SSL training.

\textcolor{purple}{\textbf{The case of Fixmatch.}} \label{RelatedWorkFixMatch}
FixMatch \citep{FixMatch} is the main focus of this paper and is one of the simplest and most effective methods in the line of SSL algorithms. FixMatch simplifies previous work in SSL by unifying pseudo-labeling methods and strong augmentation methods. In particular, for each sample and for each iteration, FixMatch produces a pseudo-label for a weak augmentation of the unlabeled sample if it crosses a fixed threshold; it then utilizes the pseudo-label as the label for a strong augmentation of the sample unlabeled sample. 
In doing so, the authors significantly simplified existing SSL methods and achieve state-of-the-art performance on a number of benchmarks such as 94.93\% on CIFAR10 with only 250 labeled samples and 88.61\% accuracy with only 40. 

\textcolor{purple}{\textbf{Other SSL techniques.}}
There are also many other areas of SSL \citep{joachim1999trans, zhu2003semi, bengio2006label, salak2007deepbelief, kingma2014semi, salimans2016improved, odena2016semi, chen2020negative} and related areas of self-supervised learning \citep{ting2020simclr, he2019moco, ting2020big, chen2020mocov2, tian2019contrastive}.

\begin{figure*}
    \centering
    \begin{subfigure}{\linewidth}
        \includegraphics[width=\linewidth]{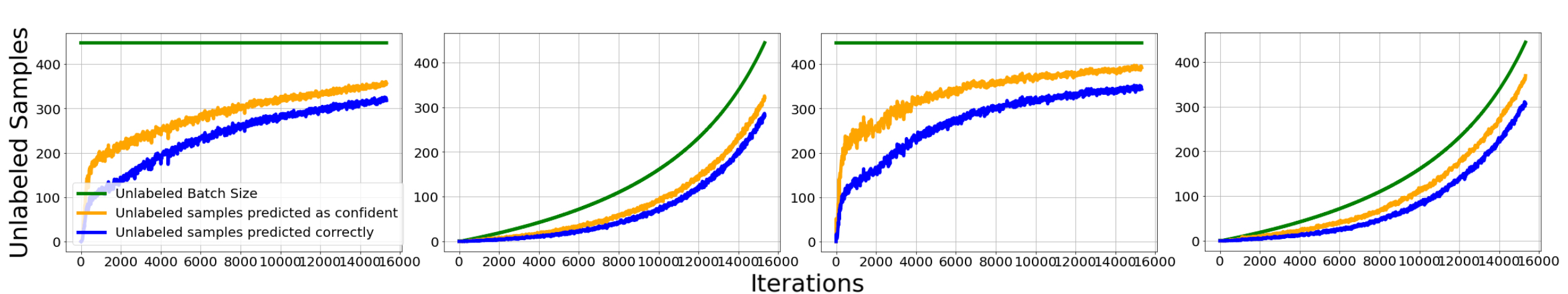}
    \end{subfigure}
    \caption{Example of data utilization of FixMatch during training measured on the batch level, averaged over the last 10 iterations. The green curve (``\textbf{\textcolor{purple}{---}}'') indicates maximum unlabeled data utilization. The orange curve (``\textbf{\textcolor{orange}{---}}'') indicates the current unlabeled data utilization of the batch. The blue curve (``\textbf{\textcolor{blue}{---}}'') indicates the amount of the confident unlabeled samples which are predicted correctly. Vanilla FixMatch utilizes a fixed batch size $=$ flat green line. Alternatively, Curriculum Batch Size alters the unlabeled batch size during training which results in an increasing green curve. Plotted above is CIFAR10 with all but 250 labels removed. Left: FixMatch \citep{FixMatch}. Middle Left: Curriculum Pseudo Labeling \citep{FlexMatch}. Middle Right: Curriculum Batch Size. Right: Curriculum Pseudo Labeling + Curriculum Batch Size. 
    }
    \label{fig:FixMatchDataUtil}
\end{figure*}



\vspace{-0.2cm}
\subsection{Curriculum Learning}
\vspace{-0.2cm}
Curriculum learning improves the performance of deep learning training by ordering training data using certain schemes into a ``curriculum'' \citep{CurriculumLearning1, CurriculumLearning2}.
Typically, this is achieved by presenting easy samples first and hard samples later. 
Much progress has been made in this field on the optimal definition of ``easy'' and ``hard'' \citep{OnThePower, LabelSimilarityCurriculum, CurriculumRevisiting, CurriculumBySmoothing, Superloss}.  
We introduce Curriculum Learning based on whether the curriculum depends on the loss, label, feature space, or is fixed or entirely learnable.

\textcolor{purple}{\textbf{Loss-based Curricula.}}
In \citep{OnThePower}, the authors use confidence of the teacher to order training data, where the teacher is either a pre-trained teacher network, or the student itself trained on data without a curriculum.
\citep{CurriculumRevisiting} revisits Pseudo-Labeling in SSL by devising a train and re-train curriculum, each time taking the top $x$\% of most confident pseudolabels from 0 to 100 in increments of 20.

\textcolor{purple}{\textbf{Label-based Curricula.}}
In \citep{FlexMatch}, the authors propose a curriculum threshold for FixMatch, which increases data utilization by lowering the threshold resulting in more pseudolabels.
To address training on imbalanced data, \citep{dynamicCurriculum} proposes a curriculum which downsamples or upsamples --depending on majority or minority classes--- and a parameter which balances between cross entropy loss and triplet loss.

\textcolor{purple}{\textbf{Feature Space-based Curricula.}}
CurriculumNet \citep{CurriculumNet} produces a curriculum based on feature space density. The authors train a model, compute embeddings, then retrain the model from clean samples to noisy samples, where samples with few nearby embeddings are more noisy.
Instead of using one-hot labels, \citep{LabelSimilarityCurriculum} improves performance by defining a probability distribution over the classes based on class similarity given by inner product similarity of items of each class.

\textcolor{purple}{\textbf{Fixed-Curricula.}}
For Adaptive Curriculum Learning \citep{AdaptiveCurriculum}, the authors propose an Exponential Moving Average of the loss in addition to a bag of tricks.
\citep{SelfPacedMedical} applies Contrastive Learning to SSL in the domain of medical images by adding a weight to the contrastive loss which decreases over time.
For noisy labels, \citep{RobustCurriculumLearning} proposes to learn from clean data first then noisy data with pseudolabels from a time-ensemble of model and data augmentations.
In \citep{CurriculumBySmoothing}, the authors add gaussian noise which decreases over time.

\textcolor{purple}{\textbf{Learnable Curricula.}}
\citep{DataParameters} adds a learnable parameter for each sample and class which scales the logits and uses backpropagation to update the learnable parameter.
\citep{SelfPacedCNN} also adds a hyperparameter for Curriculum Learning which can be optimized through backpropagation for Convolutional Neural Networks.
\citep{Superloss} proposes a ``SuperLoss" which automatically downweights samples with a large loss, i.e. ``harder" samples.


\section{The idea of Curriculum Batch Size} \label{ResultsComputationDefinition}
\textcolor{purple}{\textbf{What is the issue with FixMatch?}}
Within each batch of FixMatch \citep{FixMatch}, there are labeled samples --which FixMatch treats regularly with the cross entropy loss-- and there are unlabeled samples.
For the latter, FixMatch predicts the label based on the weak augmentation, and if it crosses a threshold, FixMatch uses this label during the strong augmentation phase. 
Since FixMatch uses a fixed batch size of 64 for labeled samples and 448 for unlabeled samples, this results in a maximum of $64 + 448 + 448 = 960$ forward passes and $64 + 448 = 512$ backwards passes per minibatch. 

For cases such as CIFAR-10 with all but 250 labels, FixMatch spends most of the training procedure above 80\% test accuracy, which means it spends close to the maximum number of forwards and backwards passes. For that setting, as shown in Table \ref{FastFixMatchTable} later on, FixMatch requires $2^{20}$ iterations (as in \cite{FixMatch}) to reach the cited error rate, which approximately equals to $2^{20} \cdot (0.5 \cdot (64 + 448 + 448) + 0.5 \cdot (64 + 448)) \approx 770$M \text{ forwards and backwards passes.}
In sequence, this implies $\approx 770\text{M} / 50000 = 15400 \text{ training epochs}$, since there are 50K training samples in CIFAR-10. 
\textit{The computational requirements are high compared to CIFAR-10 in the supervised learning scenario \cite{he2016deep}.}

\textcolor{purple}{\textbf{The Data Utilization Issue.}}
One of the main challenges of FixMatch lies in data utilization, where the number of samples contributing to the loss is limited by the choice of threshold, often fixed at 0.95. 
\textit{Data utilization is defined as the percentage of unlabeled samples above the threshold on either mini-batch level or across the whole training.}

In Figure \ref{fig:FixMatchDataUtil}, we plot the maximum unlabeled data utilization (green curve), the current unlabeled data utilization of the batch (orange curve), and the unlabeled data predicted correctly (blue curve), on the batch level. To provide some perspective: $i)$ \textit{An orange curve that follows closely with the green curve indicates high unlabeled data utilization}; $ii)$ \textit{A blue curve that closely follows the orange curve indicates an accurate model.}

As a note here, our proposal --Curriculum Batch Size or CBS-- increases the unlabeled data utilization ratio, given by the much closer green and orange curves. 
For vanilla FixMatch (leftmost plot), there is a large gap between the green and orange curves, particularly in the beginning of training but even late in training.

\textcolor{purple}{\textbf{Using Curriculum Pseudo Labeling.}}
The authors of \citep{FlexMatch} proposed a Curriculum Pseudo Labeling (CPL) , which applies a curriculum threshold. 
The intuition behind CPL is as follows: early in training, the model is often not confident enough to reach the pre-defined threshold; however, predictions may still be accurate enough to aid training. 
The authors report improved final accuracy on a range of Computer Vision benchmarks including CIFAR10/100, SVHN, STL10 and ImageNet, by applying the curriculum to FixMatch in comparison to vanilla FixMatch.
However, there is still room for improvements in data utilization (see Table \ref{CIFAR10250DataUtilTable}). 
In particular, further optimizations can be made to exploit the natural progression of training where the model performs worse at the beginning of training and better at the end of training

\subsection{Curriculum Batch Size}
The Curriculum Batch Size (or CBS) is motivated both by the observation of low data utilization and model performance progression. 
In a nutshell, CBS starts with a small unlabeled batch size and progressively increases the batch size following a curriculum.  

In particular, let the labeled samples be given by $\{(x_i, y_i)\}_{i = 1}^L$. 
The unlabeled samples are given by $\{x_i\}_{i = 1}^U$.
Following ordinary training implementations, on iteration $t$, we select the next $l_t$ labeled samples and $u_t$ unlabeled samples. 
Here, $l_t$ is fixed as $l_t := l = 64$. 
For FixMatch, $u_t := u = 448$.

\begin{figure}
    \centering
    \begin{subfigure}{\linewidth}
        \includegraphics[width=\linewidth]{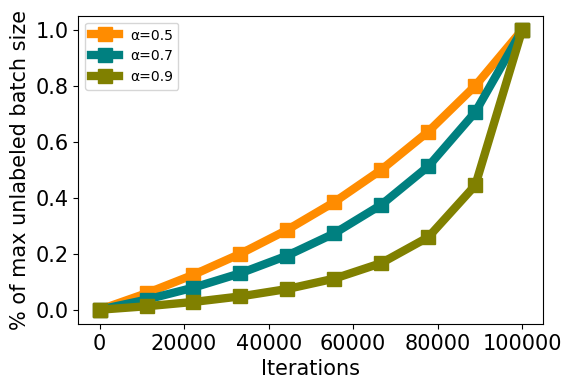}
    \end{subfigure}
    \caption{B-EXP curve for unlabeled Curriculum Batch Size depending on $\alpha$. Higher $\alpha$ is more extreme.}
    \label{fig:CurriculumShapes}
\end{figure}

\textcolor{purple}{\textbf{Schedule proposal.}} In CBS, we propose the following schedule for $u_t = \text{B-EXP}(u, t, T)$. 
Here, \text{B-EXP} stands for Bounded Exponential formulation, as shown in Figure \ref{fig:CurriculumShapes}, which allows for a smooth increase in batch size:
\begin{equation}
    \text{B-EXP} = u \cdot \Bigg( 1 -  \frac{1 - \tfrac{t}{T}}{(1 - \alpha) + \alpha \cdot \left(1 - \tfrac{t}{T}\right)} \Bigg), ~~\alpha=0.7.\nonumber
\end{equation}
Here, $u$ is the original (or maximum) unlabeled batch size; $\alpha$ is a fixed parameter; $t$ is the current iteration; and $T$ is the total iterations.
We set the parameter $\alpha=0.7$ to fix the shape of the unlabeled batch size curriculum. 
This curve was initially proposed in learning rate schedules \citep{REX, li2020budgeted}. 

In addition, following the scaling of linear learning rates \citep{LinScaleRule}, we scale the $\lambda$ coefficient of the unsupervised loss linearly with respect to the ratio of unlabeled batch size to labeled batch size. 
For example, if the current unlabeled batch size is 96 and the labeled batch size is 64, we use $\lambda = 96/64 = 1.5$. Thus, since the unlabeled batch size follows the curriculum, the $\lambda$ coefficient also follows the curriculum.

\textcolor{purple}{\textbf{Algorithm.}} Fast FixMatch is a combination of Curriculum Batch Size, labeled strong augmentation, and Curriculum Pseudo Labeling, given in Algorithm 1. 
These three methods have a certain synergy that performs better than a sum of parts, explained later in this paper. 
We also perform an ablation study further to understand the contributions, of which Curriculum Batch Size is a significant contributor.

\begin{algorithm*}
    \centering
    \caption{\textsc{$\text{Fast FixMatch}$}}\label{alg:fast_fixmatch}
    \begin{algorithmic}[1]
        \STATE \textbf{Input}: Labeled mini batch size $l$. Unlabeled mini batch size $u$. Maximum threshold $\texttt{Th}$. Total training steps $N$. Total classes $C$. Labeled training data $\{(x_i, y_i)\}_{i = 1}^L$. Unlabeled training data $\{x_i\}_{i = 1}^U$. 
        \STATE $\hat{u}_i = -1 : i\in[U]$ \hfill \texttt{// Model predictions for CPL}
        \FOR{$T = 1, \dots, T$} 
        \FOR{$c = 1, \dots, C$} 
        \STATE $T_c = \text{CPL}(c, \hat{u})$ \hfill \texttt{// Dynamic threshold according to CPL}
        \ENDFOR
        \STATE $l_t = l$ \hfill \texttt{// Labeled batch size is fixed}
        \STATE $u_t = \text{B-EXP}(u, t, T)$ \hfill \texttt{// Curriculum Batch Size for unlabeled batch size}
        \STATE $X_l, X_u \leftarrow \text{next } l_t \text{ labeled samples and } u_t \text{ unlabeled samples}$
        \STATE Apply FixMatch with strong labeled augmentation($X_l$, $X_u$, $\texttt{Th}$)
        \FOR{$c = 1, \dots, C$} 
        \STATE Update $\hat{u}_i$.
        \ENDFOR
        \ENDFOR
    \end{algorithmic}
\end{algorithm*}

\begin{table*}[!h]
\setlength\tabcolsep{3pt}
    \centering
    \small   
    \caption{CIFAR10 with all but 250 labels removed. \textcolor{lime}{Line in lime} is vanilla FixMatch; \textcolor{yellow}{Light gray line} is FixMatch with CBS; \textcolor{lightgray}{Light gray line} is FixMatch with strong augmentation; \textcolor{cyan}{Cyan line} is FixMatch with CPL; \textcolor{orange}{Orange line is Fast FixMatch}. Epoch is defined as 50,000 forward and backwards passes. Each entry is the number of total pre-defined epochs to reach a particular accuracy, and the computational decrease multiplier compared to vanilla FixMatch. Lower epochs is better.}
    \begin{tabular}{clclclclcccccccc} 
    \toprule
        Unlabeled &&  Curriculum && Labeled && Curriculum && \multicolumn{8}{c}{\multirow{2}{*}{Epochs to X\% accuracy$_{(\text{speedup over FixMatch})}$}} \\ 
        Batch && Batch && Strong && Pseudo && \multicolumn{8}{c}{} \\ 
        Size && Size && Aug && Labeling && 30\% & 40\% & 50\% & 60\% & 70\% & 80\% & 85\% & 90\% \\ 
        \cmidrule{1-1} \cmidrule{3-3} \cmidrule{5-5} \cmidrule{7-7} \cmidrule{9-16}
        \rowcolor{lime}
        448 && - && - && - && 14 & 24 & 39 & 61 & 98 & 187 & 324 & 678 \\  [0.1cm]
        \rowcolor{yellow}
        448 && \checkmark && - && - && $\phantom{0}5_{\textbf{(2.8\text{x})}}$ & $\phantom{0}8_{\textbf{(3.0\text{x})}}$ & $15_{\textbf{(2.6\text{x})}}$ & $27_{\textbf{(2.3\text{x})}}$ & $48_{\textbf{(2.0\text{x})}}$ & $105_{\textbf{(1.8\text{x})}}$ & $233_{\textbf{(1.4\text{x})}}$ & $640_{\textbf{(1.1\text{x})}}$ \\  [0.1cm]
        \rowcolor{lightgray}
        448 && - && \checkmark && - && $10_{\textbf{(1.4\text{x})}}$ & $17_{\textbf{(1.4\text{x})}}$ & $27_{\textbf{(1.4\text{x})}}$ & $39_{\textbf{(1.6\text{x})}}$ & $69_{\textbf{(1.4\text{x})}}$ & $150_{\textbf{(1.2\text{x})}}$ & $263_{\textbf{(1.2\text{x})}}$ & $678_{\textbf{(1.0\text{x})}}$  \\  [0.1cm]
        \rowcolor{cyan}
        448 && - && - && \checkmark && $12_{\textbf{(1.2\text{x})}}$ & $18_{\textbf{(1.3\text{x})}}$ & $32_{\textbf{(1.2\text{x})}}$ & $57_{\textbf{(1.1\text{x})}}$ & $90_{\textbf{(1.1\text{x})}}$ & $166_{\textbf{(1.1\text{x})}}$  & $294_{\textbf{(1.1\text{x})}}$ & $577_{\textbf{(1.2\text{x})}}$ \\  [0.1cm]
        448 && \checkmark && \checkmark && - && $\phantom{0}4_{\textbf{(3.5\text{x})}}$ & $\phantom{0}6_{\textbf{(4.0\text{x})}}$ & $\phantom{0}9_{\textbf{(4.3\text{x})}}$ & $13_{\textbf{(4.7\text{x})}}$ & $27_{\textbf{(3.6\text{x})}}$ & $\phantom{0}70_{\textbf{(2.7\text{x})}}$ & $137_{\textbf{(2.4\text{x})}}$  & $317_{\textbf{(2.1\text{x})}}$ \\   [0.1cm]
        448 && - && \checkmark && \checkmark && $12_{\textbf{(1.2\text{x})}}$ & $18_{\textbf{(1.3\text{x})}}$ & $25_{\textbf{(1.6\text{x})}}$ & $36_{\textbf{(1.7\text{x})}}$ & $63_{\textbf{(1.6\text{x})}}$ & $136_{\textbf{(1.4\text{x})}}$ & $248_{\textbf{(1.3\text{x})}}$ & $553_{\textbf{(1.2\text{x})}}$  \\  [0.1cm]
        448 && \checkmark && - && \checkmark && $\phantom{0}4_{\textbf{(3.5\text{x})}}$ & $\phantom{0}8_{\textbf{(3.0\text{x})}}$ & $15_{\textbf{(2.6\text{x})}}$ & $26_{\textbf{(2.3\text{x})}}$ & $43_{\textbf{(2.3\text{x})}}$ & $\phantom{0}97_{\textbf{(1.9\text{x})}}$ & $203_{\textbf{(1.6\text{x})}}$ & $434_{\textbf{(1.6\text{x})}}$  \\  [0.1cm]
        \rowcolor{orange}
        448 && \checkmark && \checkmark && \checkmark && $\phantom{0}5_{\textbf{(2.8\text{x})}}$ & $\phantom{0}7_{\textbf{(3.4\text{x})}}$ & $\phantom{0}10_{\textbf{(3.9\text{x})}}$ & $13_{\textbf{(4.7\text{x})}}$ & $22_{\textbf{(4.5\text{x})}}$ & $\phantom{0}62_{\textbf{(3.0\text{x})}}$ & $121_{\textbf{(2.7\text{x})}}$ & $282_{\textbf{(2.4\text{x})}}$  \\  [0.1cm] \midrule
         \multicolumn{8}{l}{\textbf{Overall speedup}} & \textbf{2.8x} & \textbf{3.4x} & \textbf{3.9x} & \textbf{4.7x} & \textbf{4.5x} & \textbf{3.0x} & \textbf{2.7x} & \textbf{2.4x}  \\
        \bottomrule
    \end{tabular}
     
    \label{CIFAR10250FastFixMatch}
\end{table*}

\begin{table*}[!h]
    \centering    
    \caption{Epochs to attain error rates reported by FixMatch \citep{FixMatch} for each number of labeled samples. For each dataset, one epoch is $X$ forward or backwards passes where $X/2=~$number of labeled + unlabeled samples.$^\dagger$ Lower epochs is better.
    }
    \setlength\tabcolsep{4pt}
    \begin{tabular}{clccclccclccclc} 
    \toprule
         & & \multicolumn{3}{c}{CIFAR-10} & & \multicolumn{3}{c}{CIFAR-100} & & \multicolumn{3}{c}{SVHN} & & STL-10 \\\cmidrule{1-1} \cmidrule{3-5} \cmidrule{7-9} \cmidrule{11-13} \cmidrule{15-15} 
         Labeled samples && 40 & 250 & 4000 && 400 & 2500 & 10000 && 40 & 250 & 1000 && 1000 \\
         Error target \citep{FixMatch} && 13.81 & 5.07 & 4.26 && 48.85 & 28.29 & 22.60 && 3.96 & 2.48 & 2.28 && 7.98 \\
         \cmidrule{1-1} \cmidrule{3-5} \cmidrule{7-9} \cmidrule{11-13} \cmidrule{15-15} 
         \rowcolor{lime}
         FixMatch (epochs) && 3300 & 15000 & 3000 && 3015 & 3768 & 3466 && 4026 & 3820 & 3924 && 2015  \\ 
         \rowcolor{orange}
         Fast FixMatch (epochs) && 977 & 7054 & 1357 && 1248 & 1357 & 1190 && $\phantom{0}781^{\star}$ & $\phantom{0}1171^{\star}$ & $\phantom{0}1152^{\star}$ && 801  \\\midrule
         \textbf{Overall speedup} && \textbf{3.4x} & \textbf{2.1x} & \textbf{2.2x} && \textbf{2.4x} & \textbf{2.8x} & \textbf{2.6x} && \textbf{5.2x} & \textbf{3.3x} & \textbf{3.4x} && \textbf{2.5x}  \\
        \bottomrule
    \end{tabular}
    \label{FastFixMatchTable} 
    \begin{scriptsize}
    \begin{flushleft}
    $^\dagger$ For example, 50,000 forward passes and 50,000 backwards passes is 1 epoch for CIFAR-10. Alternatively, 75,000 forward passes and 25,000 backwards passes is also 1 epoch for CIFAR-10. 
    \end{flushleft}
    \end{scriptsize} \vspace{-0.5cm}
    \begin{scriptsize}
    \begin{flushleft}
    $^\star$      Since Curriculum Pseudo Labeling is documented to worsen SVHN performance \citep{FlexMatch}, we remove it from Fast FixMatch for SVHN. We note that significant speedup can be achieved without Curriculum Pseudo Labeling in other settings (See Table \ref{CIFAR10250FastFixMatch})
    \end{flushleft}
    \end{scriptsize}
\end{table*}

\begin{table*}
    \centering
    \caption{CIFAR10 with all but 250 labels removed. SVHN with 250 labels and extra dataset as unlabeled. Data utilization of unlabeled data of different methods trained for 225 epochs measured across the whole training cycle. Higher data utilization is better.}
    \begin{tabular}{clclc} 
    \toprule
        Method && CIFAR10 Data Utilization && SVHN Data Utilization\\ 
        \cmidrule{1-1} \cmidrule{3-3} \cmidrule{5-5} 
        FixMatch \citep{FixMatch} && 63.6\% && 49.5\% \\ 
        Curriculum Pseudo Labeling \citep{FlexMatch} && 75.0\% && 61.1\% \\ 
        Curriculum Batch Size && 69.0\% && 59.2\% \\ 
        Curriculum Batch Size + Pseudo Labeling &&  \textbf{78.7\%} && \textbf{72.1\%}\\ 
        \bottomrule
    \end{tabular}
    
    \label{CIFAR10250DataUtilTable}
\end{table*}

\begin{table*}[!h]
    \centering
    \small
    \caption{CIFAR10 with all but 4000 labels removed. Comparing the average batch size reduction of different $\alpha$ with labeled batch size of 64 and ratio of unlabeled/labeled data samples within batch $\mu=7$. Each method run for 1357 epochs.}
    \begin{tabular}{clclclc} 
    \toprule
        Method && Average unlabeled batch size as \% && Overall average batch size as \% && Error \\ 
        \cmidrule{1-1} \cmidrule{3-3} \cmidrule{5-5} \cmidrule{7-7}
        Fast FixMatch ($\alpha=0.5$) && 38.6\% && 40.0\% && 4.57\% \\
        Fast FixMatch ($\alpha=0.7$) && 30.9\% && 36.8\% && \textbf{4.18\%} \\
        Fast FixMatch ($\alpha=0.9$) && 17.3\% && 24.5\% && 4.58\%\\
        \bottomrule
    \end{tabular}
    \label{CIFAR10250ShapesTable}
\end{table*}

\section{Results in Standard SSL Scenario}

\textcolor{purple}{\textbf{Training details.}} We reproduce the settings from FixMatch \citep{FixMatch} and utilize the same hyperparameters. For Fast FixMatch, we use the exact same hyperparameters as the original FixMatch hyperparameters, and further improvements may be possible with further tuning. For Curriculum Pseudo Labeling, we use the convex formulation which is the formulation used for the mainline results in the paper \citep{FlexMatch}. For strong labeled augmentation, we use AutoAugment \citep{cubuk2018autoaugment} for CIFAR-10, CIFAR-100 and SVHN, and RandAugment \citep{cubuk2020randaugment} for STL-10. We use WRN-28-2 for CIFAR-10 and SVHN, WRN-28-8 for CIFAR-100, and WRN-37-2 for STL-10.

In our experiments, we use the same notation with FixMatch \citep{FixMatch}, where $\mu$ is defined as the hyperparameter that determines the relative sizes of the labeled and unlabeled samples within a batch. 
In particular, when, e.g., $\mu = 2$, this indicates that the batch includes twice as many unlabeled examples, as compared to the labeled examples within the batch.

\textcolor{purple}{\textbf{Ablation study.}} Fast FixMatch uses three components: Curriculum Batch Size, labeled strong augmentation, and Curriculum Pseudo Labeling. 
We performed an ablation to understand this synergy; see Table \ref{CIFAR10250FastFixMatch}. 

On their own, \textcolor{lightgray}{labeled strong augmentation}, \textcolor{cyan}{Curriculum Pseudo Labeling} and \textcolor{yellow}{Curriculum Batch Size} all reduce computation for larger error targets, but the improvements diminish for smaller error targets. 
Labeled strong augmentation + Curriculum Pseudo Labeling reduces computation for larger error targets, but again do not reduce computation for smaller error targets. 
Both Curriculum Batch Size + labeled strong augmentation and Curriculum Batch Size + Pseudo Labeling produce substantial computational reduction for both smaller and larger error targets. 
In particular, Curriculum Batch Size + labeled strong augmentation is the best two method combination. 
Finally, \textcolor{orange}{combining all three methods improve over any one or two combinations; this is Fast MixMatch}. 
Furthermore, \textit{the improvement is more than naively multiplying the individual improvements which leads to a synergy that justifies Fast FixMatch as a combination of all three}, with Curriculum Batch Size as the main contributor.

\textcolor{purple}{\textbf{Acceleration Results.}}
We reproduce the results of FixMatch \citep{FixMatch} across the CIFAR-10, CIFAR-100, SVHN and STL-10 settings, given in Table \ref{FastFixMatchTable}. The table shows the number of pre-defined epochs required to train the model to the cited error rate. Across all different settings, Fast FixMatch achieves between $2\times-3.5\times$ speedup in the number of computations, given by the definition in Section \ref{ResultsComputationDefinition}. 
Since Fast FixMatch uses an average overall batch size of about a third of FixMatch (see also Table \ref{CIFAR10250ShapesTable}), this means that Fast FixMatch and FixMatch take roughly the same number of iterations, but the batch size for Fast FixMatch is significantly smaller. 
These gains indicate that Fast FixMatch can be efficiently used in smaller GPUs or on the edge where smaller batch sizes directly lead to wall clock speedup. 

\begin{figure}[!h]
\centering
        \includegraphics[width=0.8\linewidth]{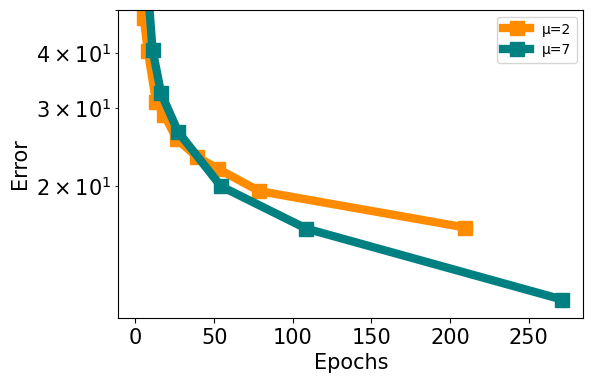}
        \includegraphics[width=0.8\linewidth]{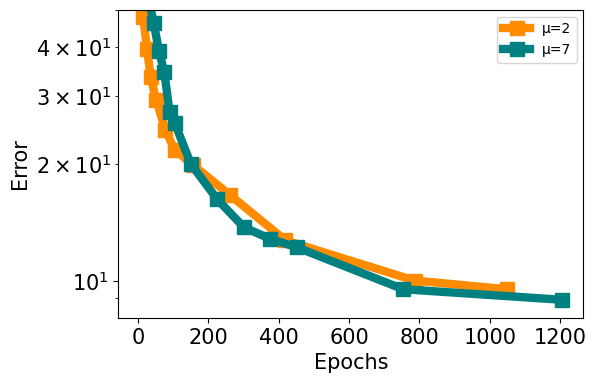}
    \caption{CIFAR10 with all but 250 labels removed. Testing if naively reducing the unlabeled batch size can result in the same gains as Fast FixMatch. Top: Comparing FixMatch with $\mu=7$ (unlabeled batch size $=448$) and $\mu=2$ (unlabeled batch size $=128$). Bottom: Comparing Fast FixMatch with $\mu=7$ (unlabeled batch size $=448$) and $\mu=2$ (unlabeled batch size $=128$).}
    \label{fig:FixMatchMu7Vs2}
\end{figure}

Using a smaller batch size naively does not result in the improvements of Fast FixMatch over FixMatch (see Figure \ref{fig:FixMatchMu7Vs2}). A smaller batch size decreases compute initially for larger error targets, but the improvements disappear for smaller error targets.

\textcolor{purple}{\textbf{Data Utilization Results.}} 
In Table \ref{CIFAR10250DataUtilTable}, we show the data utilization calculated across the entire training cycle. Curriculum Batch Size significantly increases data utilization over FixMatch and directly synergizes with Curriculum Pseudo Labeling. Curriculum Batch Size + Pseudo Labeling achieves by far the highest data utilization, further justifying the inclusion of both in Fast FixMatch. At 78.7\% for CIFAR-10, this means that 78.7\% of unlabeled samples seen during training are above the threshold and contribute to the loss. Namely, Curriculum Batch Size + Pseudo Labeling improves 15.1\% data utilization over FixMatch and 3.7\% over either Curriculum Batch Size or Pseudo Labeling for CIFAR-10, and 22.6\% data utilization over FixMatch and 11.0\% over either Curriculum Batch Size or Pseudo Labeling for SVHN.

        
    

\textcolor{purple}{\textbf{Unlabeled Batch Size Curriculums}} Since we fix the labeled batch size, there are diminishing returns for more extreme unlabeled batch size curriculums. In Table \ref{CIFAR10250ShapesTable}, we see that larger $\alpha$ results in more extreme unlabeled batch size curriculum and that $\alpha=0.7$ is a sweet spot. The curvature of these $\alpha$ are displayed in Figure \ref{fig:CurriculumShapes}.

\section{Results in Federated Self Supervised Learning}

\textcolor{purple}{\textbf{Background on Federated Self Supervised Learning}}.
Federated Learning (FL) \cite{mcmahan2017communication, https://doi.org/10.48550/arxiv.1812.06127, https://doi.org/10.48550/arxiv.1910.06378} is a distributed learning protocol that has witnessed fast development the past demi-decade. 
FL deviates from the traditional distributed learning paradigms and 
allows the integration of edge devices ---such as smartphones \cite{stojkovic2022applied}, drones \cite{qu2021decentralized}, and IoT devices \cite{nguyen2021federated}--- in the learning procedure.
However, in real life federated learning scenario, user data on local device might be largely unlabeled (such as picture with no caption). Therefore, in order to better utilize such unlabeled data, federated self-supervised learning has recently been one of the focus of SSL research and federated learning \cite{makhija2022federated,zhuang2022divergence}. 

\textcolor{purple}{\textbf{Federated Self Supervised Learning Formulation}}. Let $S$ be the total number of clients in a distributed FL scenario.
Each client $i$ has its own local labeled data $\mathcal{D}^l_i$ and unlabeled data $\mathcal{D}^u_i$. In order to be closer to real life heterogeneous data across devices, we assume non-iid in both labeled data $\mathcal{D}^l_i$ and unlabeled data $\mathcal{D}^u_i$. Unlabeled data is created such that the whole dataset satisfies $\mathcal{D} = \cup_i \mathcal{D}^u_i$, and $\mathcal{D}^u_i \cap \mathcal{D}^u_j = \emptyset, \forall i \neq j$. Labeled data $\mathcal{D}^l_i$ is random sampled subset of unlabeled data $\mathcal{D}^u_i$, which also implies $\mathcal{D}^l_i \cap \mathcal{D}^l_j = \emptyset, \forall i \neq j$   
The goal of FL is to find a global model $\mathbf{W}$ that achieves good accuracy on all data $\mathcal{D}$, by minimizing the following optimization problem: \vspace{-0.15cm}
\begin{small}
\begin{align*}
    \mathbf{W}^\star = \underset{\mathbf{W} \in \mathcal{H}}{argmin} ~\left\{\mathcal{L}(\mathbf{W}) := \tfrac{1}{S}\sum_{i = 1}^S \ell^l\left(\mathbf{W}, \mathcal{D}^l_i\right) + \ell^u\left(\mathbf{W}, \mathcal{D}^u_i\right) \right\}, \label{eq:fl_loss} \\[-15pt] \nonumber 
\end{align*}
\end{small}
where $\ell^l\left(\mathbf{W}, \mathcal{D}^l_i\right) = \tfrac{1}{|\mathcal{D}^l_i|} \sum_{\{\mathbf{x}_j, y_j\} \in \mathcal{D}^l_i} \ell\left(\mathbf{W}, \{\mathbf{x}_j, y_j\}\right)$ while $\ell^u\left(\mathbf{W}, \mathcal{D}^u_i\right) = \tfrac{1}{|\mathcal{D}^u_{i,t}|} \sum_{\{\mathbf{x}_j, \hat{y}_j\} \in \mathcal{D}^u_{i,t}} \ell\left(\mathbf{W}, \{\mathbf{x}_j, \hat{y}_j\}\right)$.
Here for labeled data we use standard supervised training loss with true label $y_j$. For unlabeled data we follow FixMatch style algorithm to select unlabeled data $\mathcal{D}^u_{i,t}$ and generate pseudo label $\hat{y}_j$ accordingly. (Here we only consider about FixMatch style SSL algorithm) 
With a slight abuse of notation, $\ell\left(\mathbf{W}, \mathcal{D}_i\right)$ denotes the total \textit{local} loss function for user $i$, associated with a local model $\mathbf{W}_i$ (not indicated above), that gets aggregated with the models of other users. 
%

\textcolor{purple}{\textbf{Fast FixMatch in Federated SSL.}} 
We extend Fast FixMatch to federated learning scenario, where each client has non-iid labeled data and non-iid unlabeled data to better represent realistic user data usage. As stated above, each client can only use its own local unlabeled data to generate pseudo label using (Fast) MixMatch and use local labeled data for supervised learning.  
In our experiments, we simulate 100 clients and sample 4 clients at every time. 
In order to ensure strong non-iid condition, we group all clients into 4 groups with strong non-iid class distributions and, at each round, we select one client from each group. 
We count the total number of labeled data in all clients. 
We use the final accuracy of Fast FixMatch as our target Accuracy to compare the speed up. 
As shown in Table \ref{Fed_FastFixMatchTable}, in both CIFAR10 and CIFAR100, Fast FixMatch achieves between 2.6 $\times$ -3.3$\times$ computation speedup. 
Since Fast FixMatch reduces the batch size, it also reduces the memory cost in memory-restricted edge devices in actual federated learning scenarios.    

\begin{table}[!h]
    \centering    
    \caption{Epochs to attain target accuracy for each number of labeled samples. For each dataset, one epoch is $X$ forward or backwards passes where $X/2=~$number of labeled + unlabeled samples.$^\dagger$ Lower epochs is better.
    }
    \setlength\tabcolsep{4pt}
    \begin{small}
    \begin{tabular}{clccclcc} 
    \toprule
         & & \multicolumn{3}{c}{CIFAR-10} & & \multicolumn{2}{c}{CIFAR-100} \\\cmidrule{1-1} \cmidrule{3-5} \cmidrule{7-8}
         Labeled samples && 160 & 1000 & 4000 &&  1000 & 4000  \\
         Accuracy target  && 56.71 & 70.64 & 71.30 &&  36.63 & 53.96  \\
         \cmidrule{1-1} \cmidrule{3-5} \cmidrule{7-8}
         \rowcolor{lime}
         FixMatch && 1101 & 990 & 970 && 734 & 825 \\ 
         \rowcolor{orange}
         FastFixMatch && 336 & 336 & 336 && 282 & 282  \\\midrule
         \textbf{Speedup} && \textbf{3.3x} & \textbf{2.9x} & \textbf{2.8x} && \textbf{2.6x} & \textbf{2.9x}  \\
        \bottomrule
    \end{tabular}
    \end{small}
    \label{Fed_FastFixMatchTable} 
\end{table}

\section{Results in Online/Streaming Learning SSL}
\textcolor{purple}{\textbf{Background in Online/Streaming Learning SSL}} Many autonomous applications would benefit from real-time, dynamic adaption of models to new data, which might come from online gathering during training. Currently, online learning1 has become a popular topic in deep learning. For instance, there are continual learning \cite{lopez2017gradient}, lifelong learning \cite{aljundi2017expert}, incremental learning \cite{rebuffi2017icarl} and streaming learning \cite{hayes2019memory}. In real life, unlabeled data may be easily collected through out the training process, which results in a streaming of new unlabeled data. Thus, Online/Streaming Learning SSL has been widely studied to fully utilize the incoming unlabeled data.

\textcolor{purple}{\textbf{Fast FixMatch in Online/Streaming Learning SSL.}}
We extend Fast Mixmatch to online/streaming learning SSL scenario with fixed small number of labeled data and streaming of unlabeled data. 
We restrict the initial set of unlabeled data to be significantly smaller (about 1/10 of original dataset) while we add new chunk of unlabeled data every 1/10 of total epochs. We use the final accuracy of Fast FixMatch as our target accuracy. As shown in Tabel \ref{Continual_FastFixMatchTable}, in both CIFAR10 and CIFAR100, Fast FixMatch achieves between 2.6 $\times$ -2.8$\times$ computation speedup. 

\begin{table}[!h]
    \centering    
    \caption{Epochs to attain target accuracy for each number of labeled samples. For each dataset, one epoch is $X$ forward or backwards passes where $X/2=~$number of labeled + unlabeled samples.$^\dagger$ Lower epochs is better.
    }
    \setlength\tabcolsep{4pt}
    \begin{small}
    \begin{tabular}{clccclccc} 
    \toprule
         &  \multicolumn{3}{c}{CIFAR-10} & \multicolumn{3}{c}{CIFAR-100} \\\midrule
         Labeled samples & 40 & 250 & 4000 &  400 & 2500 & 10000  \\
         Accuracy target  & 53.67 & 79.45 & 88.09 &  28.23 & 54.12 & 64.24  \\
         \midrule
         \rowcolor{lime}
         FixMatch & 192 & 200 & 206 & 183 & 191 & 196 \\ 
         \rowcolor{orange}
         FastFixMatch & 84 & 84 & 84 & 70 & 70 & 70 \\\midrule
         \textbf{Speedup} & \textbf{2.3x} & \textbf{2.4x} & \textbf{2.4x} & \textbf{2.6x} & \textbf{2.7x} & \textbf{2.8x} \\
        \bottomrule
    \end{tabular}
    \end{small}
    \label{Continual_FastFixMatchTable} 
\end{table}

\section{Conclusion}
In this paper, we introduce Curriculum Batch Size, a curriculum approach to batch size scaling in the SSL setting. Curriculum Batch Size uses a small batch size initially, and monotonically increases to the maximum unlabeled batch size. We use a Bounded Exponential (B-EXP) formulation to control the curriculum, and use $\alpha=0.7$ as default. We propose Fast FixMatch, a combination of Curriculum Batch Size, labeled strong augmentation, and Curriculum Pseudo Labeling. Across CIFAR-10, CIFAR-100, SVHN and STL-10 settings, we demonstrate between 2-3.5x computation improvement of Fast FixMatch over the FixMatch baseline. We perform an ablation study to understand the contribution of different components of Fast FixMatch and show that Curriculum Batch Size is a critical component, and there exists synergy better than the sum of parts. We verify that data utilization is indeed increased with Curriculum Batch Size and furthermore in combination with Curriculum Pseudo Labeling. As shown in Tabel \ref{Fed_FastFixMatchTable}, in both CIFAR10 and CIFAR100, Fast FixMatch achieves between 2.3 $\times$ -2.4$\times$ computation speed up. Finally, we extend Fast FixMatch to Federated Self-supervised Learning scenario and online/streaming learning Self-supervised Learning scenario without introducing any extra parameters/computation cost. Fast MixMatch achieves between $2.6\times$ - $3.3\times$ reduced training computations in federated SSL tasks and online/streaming learning SSL tasks, which further demonstrate the generializbility of Fast MixMatch to different scenarios and tasks.

\bibliography{FastFixmatch}

\begin{thebibliography}{66}
\providecommand{\natexlab}[1]{#1}
\providecommand{\url}[1]{\texttt{#1}}
\expandafter\ifx\csname urlstyle\endcsname\relax
  \providecommand{\doi}[1]{doi: #1}\else
  \providecommand{\doi}{doi: \begingroup \urlstyle{rm}\Url}\fi

\bibitem[Aljundi et~al.(2017)Aljundi, Chakravarty, and
  Tuytelaars]{aljundi2017expert}
R.~Aljundi, P.~Chakravarty, and T.~Tuytelaars.
\newblock Expert gate: Lifelong learning with a network of experts.
\newblock In \emph{Proceedings of the IEEE Conference on Computer Vision and
  Pattern Recognition}, pages 3366--3375, 2017.

\bibitem[Bengio et~al.(2006)Bengio, Delalleau, and Le~Roux]{bengio2006label}
Y.~Bengio, O.~Delalleau, and N.~Le~Roux.
\newblock Label propagation and quadratic criterion.
\newblock \emph{MIT Press}, 2006.

\bibitem[Bengio et~al.(2009)Bengio, Louradour, Collobert, and
  Weston]{CurriculumLearning1}
Y.~Bengio, J.~Louradour, R.~Collobert, and J.~Weston.
\newblock Curriculum learning.
\newblock In \emph{Proceedings of the 26th Annual International Conference on
  Machine Learning}, ICML '09, page 41–48, New York, NY, USA, 2009.
  Association for Computing Machinery.
\newblock ISBN 9781605585161.
\newblock \doi{10.1145/1553374.1553380}.
\newblock URL \url{https://doi.org/10.1145/1553374.1553380}.

\bibitem[Berthelot et~al.(2019)Berthelot, Carlini, Goodfellow, Papernot, and
  Raffel]{berthelot2019mixmatch}
D.~Berthelot, N.~Carlini, I.~Goodfellow, A.~Papernot, Nicolas~Oliver, and
  C.~Raffel.
\newblock Mixmatch: A holistic approach to semi-supervised learning.
\newblock \emph{arXiv preprint arXiv:1905.02249}, 2019.

\bibitem[Cascante-Bonilla et~al.(2020)Cascante-Bonilla, Tan, Qi, and
  Ordonez]{CurriculumRevisiting}
P.~Cascante-Bonilla, F.~Tan, Y.~Qi, and V.~Ordonez.
\newblock Curriculum labeling: Revisiting pseudo-labeling for semi-supervised
  learning, 2020.

\bibitem[Castells et~al.(2020)Castells, Weinzaepfel, and Revaud]{Superloss}
T.~Castells, P.~Weinzaepfel, and J.~Revaud.
\newblock Superloss: A generic loss for robust curriculum learning.
\newblock In H.~Larochelle, M.~Ranzato, R.~Hadsell, M.~Balcan, and H.~Lin,
  editors, \emph{Advances in Neural Information Processing Systems}, volume~33,
  pages 4308--4319. Curran Associates, Inc., 2020.
\newblock URL
  \url{https://proceedings.neurips.cc/paper/2020/file/2cfa8f9e50e0f510ede9d12338a5f564-Paper.pdf}.

\bibitem[Chen et~al.(2020{\natexlab{a}})Chen, Shah, and
  Kyrillidis]{chen2020negative}
J.~Chen, V.~Shah, and A.~Kyrillidis.
\newblock Negative sampling in semi-supervised learning.
\newblock \emph{ICML}, 2020{\natexlab{a}}.

\bibitem[Chen et~al.(2021)Chen, Wolfe, and Kyrillidis]{REX}
J.~Chen, C.~R. Wolfe, and A.~Kyrillidis.
\newblock {REX:} revisiting budgeted training with an improved schedule.
\newblock \emph{CoRR}, abs/2107.04197, 2021.
\newblock URL \url{https://arxiv.org/abs/2107.04197}.

\bibitem[Chen et~al.(2022)Chen, Sinha, and Kyrillidis]{chen2022stackmix}
J.~Chen, S.~Sinha, and A.~Kyrillidis.
\newblock Stackmix: A complementary mix algorithm.
\newblock In \emph{The 38th Conference on Uncertainty in Artificial
  Intelligence}, 2022.
\newblock URL \url{https://openreview.net/forum?id=HqIlPIUo5g9}.

\bibitem[Chen et~al.(2020{\natexlab{b}})Chen, Kornblith, Norouzi, and
  Hinton]{ting2020simclr}
T.~Chen, S.~Kornblith, M.~Norouzi, and G.~E. Hinton.
\newblock A simple framework for contrastive learning of visual
  representations.
\newblock \emph{CoRR}, abs/2002.05709, 2020{\natexlab{b}}.
\newblock URL \url{https://arxiv.org/abs/2002.05709}.

\bibitem[Chen et~al.(2020{\natexlab{c}})Chen, Kornblith, Swersky, Norouzi, and
  Hinton]{ting2020big}
T.~Chen, S.~Kornblith, K.~Swersky, M.~Norouzi, and G.~E. Hinton.
\newblock Big self-supervised models are strong semi-supervised learners.
\newblock \emph{CoRR}, abs/2006.10029, 2020{\natexlab{c}}.
\newblock URL \url{https://arxiv.org/abs/2006.10029}.

\bibitem[Chen et~al.(2020{\natexlab{d}})Chen, Fan, Girshick, and
  He]{chen2020mocov2}
X.~Chen, H.~Fan, R.~Girshick, and K.~He.
\newblock Improved baselines with momentum contrastive learning.
\newblock \emph{arXiv preprint arXiv:2003.04297}, 2020{\natexlab{d}}.

\bibitem[Cubuk et~al.(2018)Cubuk, Zoph, Man{\'{e}}, Vasudevan, and
  Le]{cubuk2018autoaugment}
E.~D. Cubuk, B.~Zoph, D.~Man{\'{e}}, V.~Vasudevan, and Q.~V. Le.
\newblock Autoaugment: Learning augmentation policies from data.
\newblock \emph{CoRR}, abs/1805.09501, 2018.
\newblock URL \url{http://arxiv.org/abs/1805.09501}.

\bibitem[Cubuk et~al.(2020)Cubuk, Zoph, Shlens, and Le]{cubuk2020randaugment}
E.~D. Cubuk, B.~Zoph, J.~Shlens, and Q.~Le.
\newblock Randaugment: Practical automated data augmentation with a reduced
  search space.
\newblock In H.~Larochelle, M.~Ranzato, R.~Hadsell, M.~Balcan, and H.~Lin,
  editors, \emph{Advances in Neural Information Processing Systems}, volume~33,
  pages 18613--18624. Curran Associates, Inc., 2020.
\newblock URL
  \url{https://proceedings.neurips.cc/paper/2020/file/d85b63ef0ccb114d0a3bb7b7d808028f-Paper.pdf}.

\bibitem[DeVries and Taylor(2017)]{devries2017improved}
T.~DeVries and G.~W. Taylor.
\newblock Improved regularization of convolutional neural networks with cutout,
  2017.

\bibitem[Dogan et~al.(2019)Dogan, Deshmukh, Machura, and
  Igel]{LabelSimilarityCurriculum}
{\"{U}}.~Dogan, A.~A. Deshmukh, M.~Machura, and C.~Igel.
\newblock Label-similarity curriculum learning.
\newblock \emph{CoRR}, abs/1911.06902, 2019.
\newblock URL \url{http://arxiv.org/abs/1911.06902}.

\bibitem[Elman(1993)]{CurriculumLearning2}
J.~L. Elman.
\newblock Learning and development in neural networks: the importance of
  starting small.
\newblock \emph{Cognition}, 48\penalty0 (1):\penalty0 71--99, 1993.
\newblock ISSN 0010-0277.
\newblock \doi{https://doi.org/10.1016/0010-0277(93)90058-4}.
\newblock URL
  \url{https://www.sciencedirect.com/science/article/pii/0010027793900584}.

\bibitem[Goyal et~al.(2017)Goyal, Doll{\'{a}}r, Girshick, Noordhuis,
  Wesolowski, Kyrola, Tulloch, Jia, and He]{LinScaleRule}
P.~Goyal, P.~Doll{\'{a}}r, R.~B. Girshick, P.~Noordhuis, L.~Wesolowski,
  A.~Kyrola, A.~Tulloch, Y.~Jia, and K.~He.
\newblock Accurate, large minibatch {SGD:} training imagenet in 1 hour.
\newblock \emph{CoRR}, abs/1706.02677, 2017.
\newblock URL \url{http://arxiv.org/abs/1706.02677}.

\bibitem[Grandvalet and Bengio(2005)]{grandvalet2005entmin}
Y.~Grandvalet and Y.~Bengio.
\newblock Semi-supervised learning by entropy minimization.
\newblock In \emph{Advances in Neural Information Processing Systems}, 2005.

\bibitem[Guo et~al.(2018)Guo, Huang, Zhang, Zhuang, Dong, Scott, and
  Huang]{CurriculumNet}
S.~Guo, W.~Huang, H.~Zhang, C.~Zhuang, D.~Dong, M.~R. Scott, and D.~Huang.
\newblock Curriculumnet: Weakly supervised learning from large-scale web
  images.
\newblock In \emph{European Conference on Computer Vision (ECCV)}, September
  2018.

\bibitem[Hacohen and Weinshall(2019)]{OnThePower}
G.~Hacohen and D.~Weinshall.
\newblock On the power of curriculum learning in training deep networks.
\newblock \emph{CoRR}, abs/1904.03626, 2019.
\newblock URL \url{http://arxiv.org/abs/1904.03626}.

\bibitem[Hayes et~al.(2019)Hayes, Cahill, and Kanan]{hayes2019memory}
T.~L. Hayes, N.~D. Cahill, and C.~Kanan.
\newblock Memory efficient experience replay for streaming learning.
\newblock In \emph{2019 International Conference on Robotics and Automation
  (ICRA)}, pages 9769--9776. IEEE, 2019.

\bibitem[He et~al.(2016)He, Zhang, Ren, and Sun]{he2016deep}
K.~He, X.~Zhang, S.~Ren, and J.~Sun.
\newblock Deep residual learning for image recognition.
\newblock In \emph{Proceedings of the IEEE conference on computer vision and
  pattern recognition}, pages 770--778, 2016.

\bibitem[He et~al.(2019)He, Fan, Wu, Xie, and Girshick]{he2019moco}
K.~He, H.~Fan, Y.~Wu, S.~Xie, and R.~B. Girshick.
\newblock Momentum contrast for unsupervised visual representation learning.
\newblock \emph{CoRR}, abs/1911.05722, 2019.
\newblock URL \url{http://arxiv.org/abs/1911.05722}.

\bibitem[Hendrycks et~al.(2020)Hendrycks, Mu, Cubuk, Zoph, Gilmer, and
  Lakshminarayanan]{hendrycks2020augmix}
D.~Hendrycks, N.~Mu, E.~D. Cubuk, B.~Zoph, J.~Gilmer, and B.~Lakshminarayanan.
\newblock Augmix: A simple data processing method to improve robustness and
  uncertainty, 2020.

\bibitem[Joachims(1999)]{joachim1999trans}
T.~Joachims.
\newblock Transductive inference for text classification using support vector
  machines.
\newblock In \emph{International Conference on Machine Learning}, 1999.

\bibitem[Karimireddy et~al.(2019)Karimireddy, Kale, Mohri, Reddi, Stich, and
  Suresh]{https://doi.org/10.48550/arxiv.1910.06378}
S.~P. Karimireddy, S.~Kale, M.~Mohri, S.~J. Reddi, S.~U. Stich, and A.~T.
  Suresh.
\newblock {SCAFFOLD}: Stochastic controlled averaging for federated learning,
  2019.
\newblock URL \url{https://arxiv.org/abs/1910.06378}.

\bibitem[Kingma et~al.(2014)Kingma, Mohamed, Rezende, and
  Welling]{kingma2014semi}
D.~P. Kingma, S.~Mohamed, D.~J. Rezende, and M.~Welling.
\newblock Semisupervised learning with deep generative models.
\newblock In \emph{Advances in Neural Information Processing Systems}, 2014.

\bibitem[Kong et~al.(2021)Kong, Liu, Wang, and Tao]{AdaptiveCurriculum}
Y.~Kong, L.~Liu, J.~Wang, and D.~Tao.
\newblock Adaptive curriculum learning, 2021.

\bibitem[Laine and Aila(2017)]{laine2017temporal}
S.~Laine and T.~Aila.
\newblock Temporal ensembling for semi-supervised learning.
\newblock In \emph{International Conference on Learning Representations}, 2017.

\bibitem[Li and Gong(2017)]{SelfPacedCNN}
H.~Li and M.~Gong.
\newblock Self-paced convolutional neural networks.
\newblock In \emph{Proceedings of the Twenty-Sixth International Joint
  Conference on Artificial Intelligence, {IJCAI-17}}, pages 2110--2116, 2017.
\newblock \doi{10.24963/ijcai.2017/293}.
\newblock URL \url{https://doi.org/10.24963/ijcai.2017/293}.

\bibitem[Li et~al.(2019)Li, Yumer, and Ramanan]{li2020budgeted}
M.~Li, E.~Yumer, and D.~Ramanan.
\newblock Budgeted training: Rethinking deep neural network training under
  resource constraints, 2019.
\newblock URL \url{https://arxiv.org/abs/1905.04753}.

\bibitem[Li et~al.(2018)Li, Sahu, Zaheer, Sanjabi, Talwalkar, and
  Smith]{https://doi.org/10.48550/arxiv.1812.06127}
T.~Li, A.~K. Sahu, M.~Zaheer, M.~Sanjabi, A.~Talwalkar, and V.~Smith.
\newblock Federated optimization in heterogeneous networks, 2018.
\newblock URL \url{https://arxiv.org/abs/1812.06127}.

\bibitem[Lim et~al.(2019)Lim, Kim, Kim, Kim, and Kim]{fastautoaugment}
S.~Lim, I.~Kim, T.~Kim, C.~Kim, and S.~Kim.
\newblock Fast autoaugment, 2019.
\newblock URL \url{https://arxiv.org/abs/1905.00397}.

\bibitem[Lopez-Paz and Ranzato(2017)]{lopez2017gradient}
D.~Lopez-Paz and M.~Ranzato.
\newblock Gradient episodic memory for continual learning.
\newblock \emph{Advances in neural information processing systems}, 30, 2017.

\bibitem[Makhija et~al.(2022)Makhija, Ho, and Ghosh]{makhija2022federated}
D.~Makhija, N.~Ho, and J.~Ghosh.
\newblock Federated self-supervised learning for heterogeneous clients.
\newblock \emph{arXiv preprint arXiv:2205.12493}, 2022.

\bibitem[McMahan et~al.(2017)McMahan, Moore, Ramage, Hampson, and
  y~Arcas]{mcmahan2017communication}
B.~McMahan, E.~Moore, D.~Ramage, S.~Hampson, and B.~A. y~Arcas.
\newblock Communication-efficient learning of deep networks from decentralized
  data.
\newblock In \emph{Artificial intelligence and statistics}, pages 1273--1282.
  PMLR, 2017.

\bibitem[Miyato et~al.(2017)Miyato, Maeda, Koyama, and
  Ishii]{miyato2017virtual}
T.~Miyato, S.-i. Maeda, M.~Koyama, and S.~Ishii.
\newblock Virtual adversarial training: a regularization method for supervised
  and semi-supervised learning.
\newblock \emph{arXiv preprint arXiv:1704.03976}, 2017.

\bibitem[Nguyen et~al.(2021)Nguyen, Malik, Zhan, Yousefpour, Rabbat, Malek, and
  Huba]{nguyen2021federated}
J.~Nguyen, K.~Malik, H.~Zhan, A.~Yousefpour, M.~Rabbat, M.~Malek, and D.~Huba.
\newblock Federated learning with buffered asynchronous aggregation.
\newblock \emph{arXiv preprint arXiv:2106.06639}, 2021.

\bibitem[Odena(2016)]{odena2016semi}
A.~Odena.
\newblock Semi-supervised learning with generative adversarial networks.
\newblock \emph{arXiv preprint arXiv:1606.01583}, 2016.

\bibitem[Overwijk et~al.(2022)Overwijk, Xiong, and Callan]{LargeScaleData2}
A.~Overwijk, C.~Xiong, and J.~Callan.
\newblock Clueweb22: 10 billion web documents with rich information.
\newblock In \emph{Proceedings of the 45th International ACM SIGIR Conference
  on Research and Development in Information Retrieval}, SIGIR '22, page
  3360–3362, New York, NY, USA, 2022. Association for Computing Machinery.
\newblock ISBN 978-1-4503-8732-3.
\newblock \doi{10.1145/3477495.3536321}.
\newblock URL \url{https://doi.org/10.1145/3477495.3536321}.

\bibitem[Peng et~al.(2021)Peng, Wang, Desrosiers, and
  Pedersoli]{SelfPacedMedical}
J.~Peng, P.~Wang, C.~Desrosiers, and M.~Pedersoli.
\newblock Self-paced contrastive learning for semi-supervisedmedical image
  segmentation with meta-labels.
\newblock \emph{CoRR}, abs/2107.13741, 2021.
\newblock URL \url{https://arxiv.org/abs/2107.13741}.

\bibitem[Qu et~al.(2021)Qu, Dai, Zhuang, Chen, Dong, Wu, and
  Guo]{qu2021decentralized}
Y.~Qu, H.~Dai, Y.~Zhuang, J.~Chen, C.~Dong, F.~Wu, and S.~Guo.
\newblock Decentralized federated learning for {UAV} networks: Architecture,
  challenges, and opportunities.
\newblock \emph{IEEE Network}, 35\penalty0 (6):\penalty0 156--162, 2021.

\bibitem[Rebuffi et~al.(2017)Rebuffi, Kolesnikov, Sperl, and
  Lampert]{rebuffi2017icarl}
S.-A. Rebuffi, A.~Kolesnikov, G.~Sperl, and C.~H. Lampert.
\newblock icarl: Incremental classifier and representation learning.
\newblock In \emph{Proceedings of the IEEE conference on Computer Vision and
  Pattern Recognition}, pages 2001--2010, 2017.

\bibitem[Russakovsky et~al.(2014)Russakovsky, Deng, Su, Krause, Satheesh, Ma,
  Huang, Karpathy, Khosla, Bernstein, Berg, and Li]{imagenet}
O.~Russakovsky, J.~Deng, H.~Su, J.~Krause, S.~Satheesh, S.~Ma, Z.~Huang,
  A.~Karpathy, A.~Khosla, M.~Bernstein, A.~Berg, and F.-F. Li.
\newblock Imagenet large scale visual recognition challenge.
\newblock \emph{arXiv preprint arXiv:1409.0575}, 2014.

\bibitem[Sajjadi et~al.(2016)Sajjadi, Javanmardi, and
  Tasdizen]{sajjadi2016regularization}
M.~Sajjadi, M.~Javanmardi, and T.~Tasdizen.
\newblock Regularization with stochastic transformations and perturbations for
  deep semi-supervised learning.
\newblock In \emph{Advances in Neural Information Processing Systems}, 2016.

\bibitem[Salakhutdinov and Hinton(2007)]{salak2007deepbelief}
R.~Salakhutdinov and H.~E. Hinton, Geoffrey.
\newblock Using deep belief nets to learn covariance kernels for gaussian
  processes.
\newblock In \emph{Advances in Neural Information Processing Systems}, 2007.

\bibitem[Salimans et~al.(2016)Salimans, Goodfellow, Zaremba, Cheung, Radford,
  and Chen]{salimans2016improved}
T.~Salimans, I.~Goodfellow, W.~Zaremba, V.~Cheung, A.~Radford, and X.~Chen.
\newblock Improved techniques for training gans.
\newblock In \emph{Advances in Neural Information Processing Systems}, 2016.

\bibitem[Saxena et~al.(2019)Saxena, Tuzel, and DeCoste]{DataParameters}
S.~Saxena, O.~Tuzel, and D.~DeCoste.
\newblock Data parameters: A new family of parameters for learning a
  differentiable curriculum.
\newblock In H.~Wallach, H.~Larochelle, A.~Beygelzimer, F.~d\textquotesingle
  Alch\'{e}-Buc, E.~Fox, and R.~Garnett, editors, \emph{Advances in Neural
  Information Processing Systems}, volume~32. Curran Associates, Inc., 2019.
\newblock URL
  \url{https://proceedings.neurips.cc/paper/2019/file/926ffc0ca56636b9e73c565cf994ea5a-Paper.pdf}.

\bibitem[Sinha et~al.(2020)Sinha, Garg, and Larochelle]{CurriculumBySmoothing}
S.~Sinha, A.~Garg, and H.~Larochelle.
\newblock Curriculum by texture.
\newblock \emph{CoRR}, abs/2003.01367, 2020.
\newblock URL \url{https://arxiv.org/abs/2003.01367}.

\bibitem[Sohn et~al.(2020)Sohn, Berthelot, Li, Zhang, Carlini, Cubuk, Kurakin,
  Zhang, and Raffel]{FixMatch}
K.~Sohn, D.~Berthelot, C.~Li, Z.~Zhang, N.~Carlini, E.~D. Cubuk, A.~Kurakin,
  H.~Zhang, and C.~Raffel.
\newblock Fixmatch: Simplifying semi-supervised learning with consistency and
  confidence.
\newblock \emph{CoRR}, abs/2001.07685, 2020.
\newblock URL \url{https://arxiv.org/abs/2001.07685}.

\bibitem[Stojkovic et~al.(2022)Stojkovic, Woodbridge, Fang, Cai, Petrov, Iyer,
  Huang, Yau, Kumar, Jawa, et~al.]{stojkovic2022applied}
B.~Stojkovic, J.~Woodbridge, Z.~Fang, J.~Cai, A.~Petrov, S.~Iyer, D.~Huang,
  P.~Yau, A.~S. Kumar, H.~Jawa, et~al.
\newblock Applied federated learning: Architectural design for robust and
  efficient learning in privacy aware settings.
\newblock \emph{arXiv preprint arXiv:2206.00807}, 2022.

\bibitem[Tarvainen and Valpola(2017)]{tarvainen2017mean}
A.~Tarvainen and H.~Valpola.
\newblock Mean teachers are better role models: Weight-averaged consistency
  targets improve semi-supervised deep learning results.
\newblock In \emph{Advances in Neural Information Processing Systems}, 2017.

\bibitem[Tian et~al.(2019)Tian, Krishnan, and Isola]{tian2019contrastive}
Y.~Tian, D.~Krishnan, and P.~Isola.
\newblock Contrastive multiview coding, 2019.

\bibitem[Verma et~al.(2019)Verma, Lamb, Kannala, Bengio, and
  Lopez-Pas]{verma2019ict}
V.~Verma, A.~Lamb, J.~Kannala, Y.~Bengio, and D.~Lopez-Pas.
\newblock Interpolation consistency training for semi-supervised learning.
\newblock \emph{arXiv preprint arXiv:1903.03825}, 2019.

\bibitem[Wang et~al.(2021)Wang, Wu, Pino, Baevski, Auli, and
  Conneau]{LargeScaleSSL1}
C.~Wang, A.~Wu, J.~Pino, A.~Baevski, M.~Auli, and A.~Conneau.
\newblock Large-scale self- and semi-supervised learning for speech
  translation, 2021.
\newblock URL \url{https://arxiv.org/abs/2104.06678}.

\bibitem[Wang et~al.(2019)Wang, Gan, Wu, and Yan]{dynamicCurriculum}
Y.~Wang, W.~Gan, W.~Wu, and J.~Yan.
\newblock Dynamic curriculum learning for imbalanced data classification.
\newblock \emph{CoRR}, abs/1901.06783, 2019.
\newblock URL \url{http://arxiv.org/abs/1901.06783}.

\bibitem[Xie et~al.(2019)Xie, Dai, Hovy, Luong, and Le]{xie2019unsupervised}
Q.~Xie, Z.~Dai, E.~Hovy, M.-T. Luong, and Q.~Le.
\newblock Unsupervised data augmentation for consistency training.
\newblock \emph{arXiv preprint arXiv:1904.12848}, 2019.

\bibitem[Yun et~al.(2019)Yun, Han, Oh, Chun, Choe, and Yoo]{yun2019cutmix}
S.~Yun, D.~Han, S.~J. Oh, S.~Chun, J.~Choe, and Y.~Yoo.
\newblock Cutmix: Regularization strategy to train strong classifiers with
  localizable features.
\newblock In \emph{International Conference on Computer Vision (ICCV)}, 2019.

\bibitem[Zhang et~al.(2021)Zhang, Wang, Hou, Wu, Wang, Okumura, and
  Shinozaki]{FlexMatch}
B.~Zhang, Y.~Wang, W.~Hou, H.~Wu, J.~Wang, M.~Okumura, and T.~Shinozaki.
\newblock Flexmatch: Boosting semi-supervised learning with curriculum pseudo
  labeling.
\newblock \emph{CoRR}, abs/2110.08263, 2021.
\newblock URL \url{https://arxiv.org/abs/2110.08263}.

\bibitem[Zhang et~al.(2017)Zhang, Cisse, Dauphin, and
  Lopez-Pas]{zhang2017mixup}
H.~Zhang, M.~Cisse, Y.~N. Dauphin, and D.~Lopez-Pas.
\newblock mixup: Beyond empirical risk minimization.
\newblock \emph{arXiv preprint arXiv:1710.09412}, 2017.

\bibitem[Zhang et~al.(2022)Zhang, Park, Han, Qin, Gulati, Shor, Jansen, Xu,
  Huang, Wang, Zhou, Li, Ma, Chan, Yu, Wang, Cao, Sim, Ramabhadran, Sainath,
  Beaufays, Chen, Le, Chiu, Pang, and Wu]{LargeScaleSSL2}
Y.~Zhang, D.~S. Park, W.~Han, J.~Qin, A.~Gulati, J.~Shor, A.~Jansen, Y.~Xu,
  Y.~Huang, S.~Wang, Z.~Zhou, B.~Li, M.~Ma, W.~Chan, J.~Yu, Y.~Wang, L.~Cao,
  K.~C. Sim, B.~Ramabhadran, T.~N. Sainath, F.~Beaufays, Z.~Chen, Q.~V. Le,
  C.-C. Chiu, R.~Pang, and Y.~Wu.
\newblock {BigSSL}: Exploring the frontier of large-scale semi-supervised
  learning for automatic speech recognition.
\newblock \emph{{IEEE} Journal of Selected Topics in Signal Processing},
  16\penalty0 (6):\penalty0 1519--1532, oct 2022.
\newblock \doi{10.1109/jstsp.2022.3182537}.
\newblock URL \url{https://doi.org/10.1109%2Fjstsp.2022.3182537}.

\bibitem[Zhou et~al.(2021)Zhou, Wang, and Bilmes]{RobustCurriculumLearning}
T.~Zhou, S.~Wang, and J.~Bilmes.
\newblock Robust curriculum learning: from clean label detection to noisy label
  self-correction.
\newblock In \emph{International Conference on Learning Representations}, 2021.
\newblock URL \url{https://openreview.net/forum?id=lmTWnm3coJJ}.

\bibitem[Zhu et~al.(2003)Zhu, Ghahramani, and Lafferty]{zhu2003semi}
X.~Zhu, Z.~Ghahramani, and J.~D. Lafferty.
\newblock Semi-supervised learning using gaussian fields and harmonic
  functions.
\newblock In \emph{International Conference on Machine Learning}, 2003.

\bibitem[Zhu et~al.(2015)Zhu, Kiros, Zemel, Salakhutdinov, Urtasun, Torralba,
  and Fidler]{LargeScaleData1}
Y.~Zhu, R.~Kiros, R.~Zemel, R.~Salakhutdinov, R.~Urtasun, A.~Torralba, and
  S.~Fidler.
\newblock Aligning books and movies: Towards story-like visual explanations by
  watching movies and reading books, 2015.
\newblock URL \url{https://arxiv.org/abs/1506.06724}.

\bibitem[Zhuang et~al.(2022)Zhuang, Wen, and Zhang]{zhuang2022divergence}
W.~Zhuang, Y.~Wen, and S.~Zhang.
\newblock Divergence-aware federated self-supervised learning.
\newblock \emph{arXiv preprint arXiv:2204.04385}, 2022.

\end{thebibliography}
\end{document}